\documentclass[letterpaper, 10 pt, conference]{ieeeconf}  

\IEEEoverridecommandlockouts    

\usepackage{graphics} 
\usepackage{epsfig} 
\usepackage{epstopdf}
\usepackage{forest}
\usepackage{lipsum}

\usepackage{graphicx}
\usepackage{siunitx}
\usepackage{multirow}
\usepackage{lipsum}
\usepackage{comment}
\usepackage{float}  
\usepackage[export]{adjustbox}
\usepackage[caption=false]{subfig} 
\newenvironment{rot}{\par\color{red}}{\par}

\usepackage{url}

\usepackage{tikz}
\usepackage{textcomp}
\usepackage{lipsum}

\newcommand\copyrighttext{%
  \footnotesize \textcopyright 2021 IEEE. Personal use of this material is permitted. Permission from IEEE must be obtained for all other uses, in any current or future media, including reprinting/republishing this material for advertising or promotional purposes, creating new collective works, for resale or redistribution to servers or lists, or reuse of any copyrighted component of this work in other works.}
\newcommand\copyrightnotice{%
\begin{tikzpicture}[remember picture,overlay]
\node[anchor=south,yshift=10pt,xshift=10pt] at (current page.south) {\fbox{\parbox{\dimexpr\textwidth-\fboxsep-\fboxrule\relax}{\copyrighttext}}};
\end{tikzpicture}%
}

\newcommand\whichconference{%
  \footnotesize \centering This manuscript has been accepted as a contributed paper to be presented at the 2021 32nd IEEE Intelligent Vehicles Symposium (IV 2021)\\ July 11-15, 2021, in Nagoya University, Nagoya, Japan.}
\newcommand\conferencenote{%
\begin{tikzpicture}[remember picture,overlay]
\node[anchor=north,yshift=-10pt,xshift=10pt] at (current page.north) {\fbox{\parbox{\dimexpr\textwidth-\fboxsep-\fboxrule\relax}{\whichconference}}};
\end{tikzpicture}%
}

\title{\LARGE \bf Predicting Vehicles Trajectories in Urban Scenarios with Transformer Networks and Augmented Information}
\author{A. Quintanar$^{1}$, D. Fern\'andez-Llorca$^{1,2}$, I. Parra$^{1}$, R. Izquierdo$^{1}$ and M. A. Sotelo$^{1}$ 
\thanks{$^{1}$Computer Engineering Department, Universidad de Alcal\'a, Alcal\'a de Henares, Spain.
        {\tt\small alvaro.quintanar@uah.es}}%
        \newline
\thanks{$^{2}$European Commission, Joint Research Center, Seville, Spain.}
}

\begin{document}

\maketitle

\copyrightnotice
\conferencenote

\thispagestyle{empty}
\pagestyle{empty}

\begin{abstract}
Understanding the behavior of road users is of vital importance for the development of trajectory prediction systems. In this context, the latest advances have focused on recurrent structures, establishing the social interaction between the agents involved in the scene. More recently, simpler structures have also been introduced for predicting pedestrian trajectories, based on Transformer Networks, and using positional information \cite{Giuliari2020b}. They allow the individual modelling of each agent's trajectory separately without any complex interaction terms. Our model exploits these simple structures by adding augmented data (position and heading), and adapting their use to the problem of vehicle trajectory prediction in urban scenarios in prediction horizons up to 5 seconds. In addition, a cross-performance analysis is performed between different types of scenarios, including highways, intersections and roundabouts, using recent datasets (inD, rounD, highD and INTERACTION). Our model achieves state-of-the-art results and proves to be flexible and adaptable to different types of urban contexts.



\end{abstract}

\section{Introduction} \label{sec:introduction}

Predicting road users trajectories is essential for autonomous driving. It enables path planning taking into account future states of dynamic agents, resulting in safer and more comfortable driving. It is reasonable to think that agents are affected in their behavior by traffic conditions and road structure, so any potential solution must be flexible enough to be applicable to various scene contexts. In addition, although recent approaches model the behaviors of multiple agent types within a single model (vehicles, cyclists and pedestrians) \cite{Cheng2021}, \cite{Cheng}, having specific models for each agent type simplifies the problem, and facilitates the use of simple and effective architectures, such as Transformer (TF) networks. These have been proposed for Natural Language Processing (NLP) to deal with word sequences, using attention instead of sequential processing \cite{Vaswani}. 

TF networks have been recently applied to predict pedestrians trajectories \cite{Giuliari2020b}, by using positional information. These are considered as "simple" models because each agent is modelled separately without  considering complex interactions such as social recurrent networks \cite{Deo2018} and graph neural networks \cite{Diehl2019} approaches. In this paper, we explore, for the first time, the applicability of TF networks to predict vehicles trajectories in multiple scenarios. We study the effect of augmenting the positional information with additional variables (i.e., velocity and orientation) for the context of vehicles. We evaluate the proposed model using four recent datasets (highD \cite{Krajewski2018}, inD \cite{Bock2019}, rounD \cite{Krajewski2020} and INTERACTION \cite{Zhan2019}) which include different scenarios of different complexity (intersections, roundabouts, and straight road segments). In this way, we validate and assess the flexibility and learning transferability of the TF networks when dealing with vehicle trajectory prediction in multiple urban scenarios.  




\begin{figure}[t]
    \centering
    {
    \vspace{2pt}
    \includegraphics[width=\columnwidth]{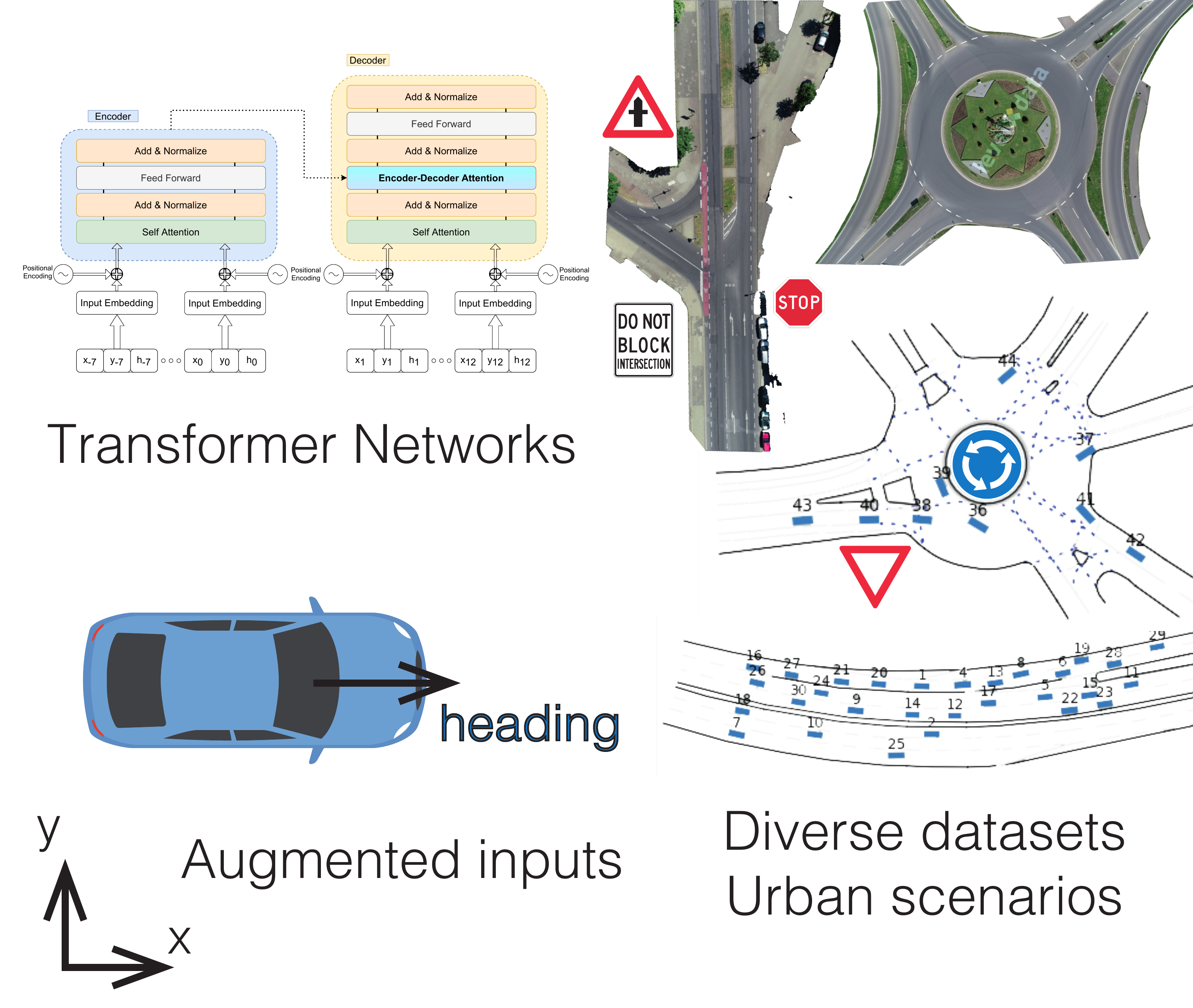}
    }
    \vspace{-20pt}
    \caption{System overview.}
    \vspace{-16pt}
    \label{fig:intro}
\end{figure}


\section{Related Work} \label{sec:sota}
In the early stages of trajectory prediction, classical approaches relied essentially on linear regression, Bayesian filtering or Markov decision process. These methods performed properly, but since they are based on a explicit physical model of the agents’ behavior, their scaling and generalization capabilities are limited. Data-driven approaches have become more predominant to address these limitations.  Recently, Deep Learning based methods have emerged for vehicle maneuvers \cite{Izquierdo2019}, \cite{biparva2021video} and trajectories \cite{Izquierdo2020} prediction. More specifically, Recurrent Neural Networks (RNNs), such as GRUs and LSTMs, have been widely used in the field. In order to account for interactions, these approaches were adapted by including a social pooling layer (Social-LSTM) for pedestrians \cite{Alahi2016} and also for vehicles \cite{Deo2018}. In order to overcome some limitations of the social pooling layer, we can find approaches based on occupancy grids \cite{Pfeiffer2018}, (Scene-LSTM) \cite{manh2019scenelstm}, message passing (SR-LSTM) \cite{Zhang2019}, Generative Adversarial Networks (Social GAN) \cite{Gupta}, (SoPhie) \cite{sohpie2018} and multi-agent tensors \cite{Zhao2019a}. 

Another interesting approach to model spatial interactions for trajectory forecast is through Graph Convolutional (GNN) or Graph Attention (GAT) Networks. They use a graph to represent each agent (nodes) and their interactions (edges), and update each node state and implement a weighted message passing mechanism by using convolutional or feed-forward layers, or attention mechanisms. They have been applied for modeling traffic participant interactions \cite{Diehl2019}. In order to integrate temporal information, graph representations are usually combined with recurrent-based ensembles such as Social-BiGAT \cite{Kosaraju2019}, Social-STGCNN \cite{Mohamed2020}, GRIP++ \cite{li2020grip}, or adapted to allow learning temporal patterns (ST-GCN) \cite{Yan}. Recently, the STAR model \cite{Yu} proposed to combine GATs to model spatial interactions, with Transformers to model temporal interactions. Finally, we can find recent proposals based on the combination of some encoder-decoder architecture with Conditional Variational Auto-Encoders (CVAE) such as AMENet \cite{Cheng2021} or DCENet \cite{Cheng}. CVAEs are used to encode spatial-temporal information into a latent space. Future trajectories of the agents are then predicted by repeatedly sampling from the learned latent space. Most of the aforementioned approaches focused on pedestrian trajectories. 

This paper is mainly inspired by \cite{Giuliari2020b} which adapted Transformer Networks (TF) to predict pedestrian trajectories in crowded spaces. They achieved state-of-the-art results in TrajNet benchmark \cite{sadeghian2018trajnet}, by relying only on self positional information without explicitly modeling interactions. TF models overcome the limitations of RNN-based models which suffer when modeling data in long temporal sequences, or in cases in which there is a lack of input data in observations (very common in real systems involving physical sensors), being more parallelizable and requiring significantly less time to train.  Moreover, its main weakness, i.e. the absence of explicit modeling of spatial interactions (which is explicitly addressed by graphical-based approaches), also represents its main strength, i.e. the simplicity of the model, which also facilitates explainability. Spatial interactions and context can be easily incorporated into the input embedding without increasing the model complexity. 

To the best of our knowledge, this is the first attempt to use TF models in the specific context of vehicle trajectory prediction. We evaluate the effect on performance of adding the heading to positional information, as well as the effect of the sampling frequency. Another important contribution is the evaluation of the system on four different datasets, which include different types of urban environments such as roundabouts, different types of intersections, straight road sections, etc. Different cross experiments are performed to validate the flexibility and generalization capability of this approach.

\vspace{-12pt}
\section{Methodology} \label{sec:methodology}

This section describes the methodology employed to deploy the model: defining input and output data, pre-processing BEV datasets and modifying a Vanilla-Transformer to introduce new inputs and process them adequately.

\subsection{Addressing the problem}
As stated before, to predict a trajectory the objective is to forecast future positions of agent $i$ by observing its current and previous positions, being defined an observation window (seen sequence) and a prediction horizon. The objective is to provide predictions about the position of the agent at the future $\kappa$ steps.

\begin{figure}[H]
    \centering
    {
    \vspace{-5pt}
    \includegraphics[width=\columnwidth]{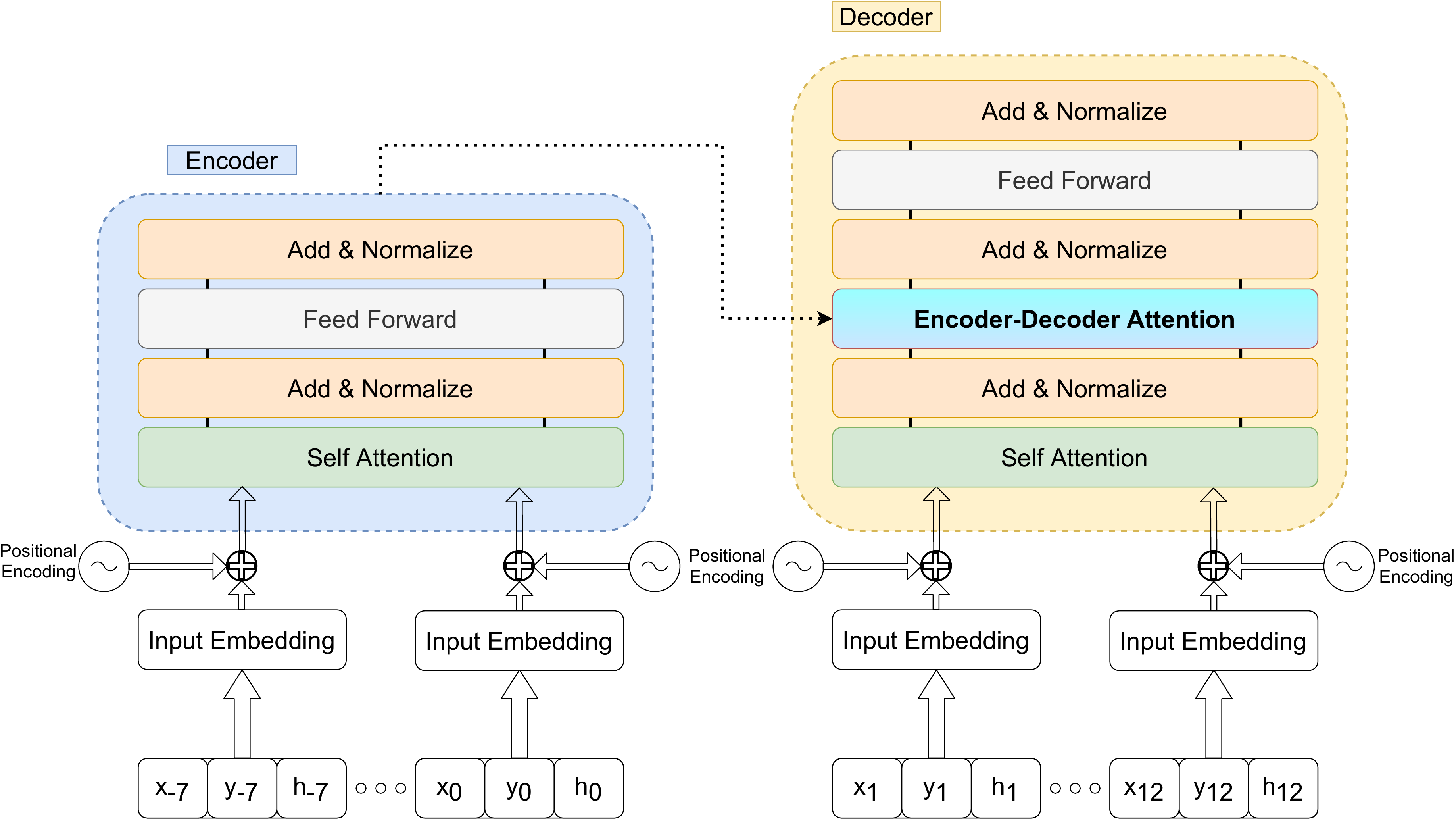}
    }
    \vspace{-10pt}
    \caption{Architecture overview: adding new inputs.}
    \vspace{-9pt}
    \label{fig:transformer}
\end{figure}

\subsubsection{Positional Information and Encoding}
Compared to an LSTM, TF does not process the input sequentially, so it needs a way to encode the temporal information. For this it makes use of positional encoding, where each input embedding has its corresponding timestamp, calculated through sine and cosine functions, as in \cite{Vaswani}. The input embedding is concatenated with the positional encoding vector.
This feature of the TF makes it possible to learn in parallel from all the time instants, while the LSTM needs to be processed sequentially, performing back-propagation. If a certain element is missing from the input, the positional encoding will consider it, which is not as problematic as in a LSTM, where this information could be lost. This claim was proven by the original authors of the model, showing that the architecture can perform satisfactorily even when missing data, degrading performance in inverse proportion to the age of the lost samples.
Normalization of the input is vital for the performance, so the inputs are normalized by subtracting the mean and dividing by SD of the train set.
This model is not multimodal, although it would be possible to classify the inputs into different classes by representing the inputs as vectors (i.e., through classification). According to its original developers, regression works better than classification-based approaches, so this approach is the one chosen.

We kept the original architecture of \cite{Giuliari2020b}, adopting an L2 loss in which position increments (to enhance the independence of each given position) and normalized heading are configured. \(d_{model}=512\), using 6 layers and 8 attention heads. Warmup have been setup to 10 epochs, using a decaying learning rate in the subsequent epochs. 

The memory of the model is based on the attention modules, where in the encoding stage a representation for the observation sequence is produced. Thus, this creates two vectors to be handed to the decoder stage, as seen in the figure \ref{fig:transformer}. TF keeps the memory separate from the decoded sequence, unlike LSTM, which keeps everything in the hidden state. For this reason, TF is known to work better with longer prediction horizon.

The additional input to the Transformer is the heading normalized between 0 and 1. This is then combined with the training loss calculation to complement the velocity (position increments) at each time instant.

\subsubsection{Preprocessing the data}

As data contained in every dataset is enclosed in a different way, as will be detailed in the corresponding section, it is necessary to preprocess them. Firstly, it is required to take into consideration the framerate, even though in this case the input data will be kept in order to enable possible studies to be carried out, considering this matter directly in the data loader.
The sets are separated by classes, according to the tests to be performed. After analyzing part of the recordings of one of the datasets, it was noticed that there were static cars parked steadily, which data could affect the result of the inference. Thus, a filtering is applied to remove this specific information from the datasets, resulting the structure $frame, track, x, y, vx, vy, heading$. Some tests have been developed with $vx$ and $vy$, but finally they have not been employed in this work. Increments of $x$ and $y$ have been calculated directly as velocity value, considering the time step. Units are expressed in SI.
The sliding window step for each valid full trajectory capture has been set to 1, in order to analyze all possible trajectories in each split.

\section{Results}\label{sec:results}

\subsection{Datasets}

The use of bird's eye view (BEV) datasets has been remarkably extended in the recent works to develop trajectory prediction systems, emphasizing the TrajNet \cite{sadeghian2018trajnet} for pedestrian trajectory prediction. This dataset offers a Challenge that constitutes a solid multi-scenario forecasting benchmark, observing 8 values of position ground-truth (3.2 seconds) and predicting the following 12 positions (4.8 seconds). All positions are given in world plane coordinates, being the 8-12 protocol a consistent fashion for diverse datasets, as explained in the following section.

Beyond purely pedestrian-based approaches, NGSIM datasets \cite{NGSIMHW101} \cite{NGSIMI80} pioneered in providing coverage of a highway area, offering information taken from cameras mounted at a skyscraper.
Other multi-agent focused datasets have been developed in the past few years, some focusing on highway scenarios, such as highD \cite{Krajewski2018} for highway vehicle trajectory prediction, offering aerial images obtained with a drone located over various locations of german \textit{Autobahn}, Vehicle labeling ensures that the error is below 10 centimeters, providing a combined total of 147 hours of drive time on more than 100,000 vehicles.
Moreover, the authors of this dataset went further with the concept, moving to urban scenarios: inD \cite{Bock2019} and rounD \cite{Krajewski2020} record different intersections and roundabouts, respectively. 
In addition to previously mentioned, the INTERACTION Dataset \cite{Zhan2019} combines all these scenarios, including ramp merging, signalized intersections and roundabouts. This dataset also provides diverse material in driving behavior, showing multiple critical maneuvers, including an accident. These are the situations that add value to a trajectory prediction solution, and should be evaluated here in a qualitative way. Table \ref{tab:datasets} offers an overview of datasets employed to develop the experiments.
Furthermore, while the use of 2D datasets taken from drones or fixed locations in bird's eye view enables a relatively simple creation and labeling process, the ultimate purpose of such datasets would be to train models that can later be ported to vehicles with onboard sensors, which can be tested in datasets like the PREVENTION Dataset \cite{Izquierdo2019}.

\begin{table*}[t]
\caption{Datasets used in this work}
\label{tab:datasets}
\centering
\begin{tabular}{c|c|c|c|c|c|c}
\textbf{Dataset} & \textbf{Country} & \textbf{Locations} & \textbf{\# Tracks} & \textbf{Road User Types} & \textbf{Data Frequency} & \textbf{Maps} \\ \hline
\textbf{inD} & Germany & urban intersections (4) & 11500 & pedestrian, bicycle, car, truck, bus & 25 Hz & yes \\ \hline
\textbf{rounD} & Germany & \begin{tabular}[c]{@{}c@{}}(sub-)urban \\ roundabouts (3)\end{tabular} & 13746 & \begin{tabular}[c]{@{}c@{}}pedestrian, bicycle, motorcycle,\\ car, van, truck, bus, trailer\end{tabular} & 25 Hz & yes \\ \hline
\textbf{highD} & Germany & highway (6) & 110000 & car, truck & 25 Hz & no \\ \hline
\textbf{INTERACTION} & \begin{tabular}[c]{@{}c@{}}USA\\ Germany\\  China\end{tabular} & \begin{tabular}[c]{@{}c@{}}roundabout (5), intersection (4), \\ highway (2)\end{tabular} & 40054 & pedestrian/bicycle, car, truck & 10 Hz & yes
\end{tabular}
\vspace{-10pt}
\end{table*}
\addtolength{\tabcolsep}{-4pt}


\subsection{Evaluation metrics}
TrajNet performance is measured in terms of Mean Average Displacement Error (MAD/ADE) and Final Average Displacement (FAD/FDE). ADE measures the aligned Euclidean distance from the prediction w.r.t. the ground truth, making an average of the error at every time step. That is, ADE reports a mean value of the general fit of the forecast in the predicted trajectory. FDE measures the Euclidean distance at the very last step, comparing the prediction to the corresponding ground truth position. 

\subsubsection{inD: Comparative results}

In order to make a fair comparison, the model execution for this section has been performed with the same data split used by the DCENet authors to carry out their quantitative analysis. This includes all types of agents, which are loaded and analyzed globally in the results without differentiation, which may affect the results of the TF models.

Beyond this aspect, the data split proposed for the table \ref{tab:comparative-table-intersections} includes in the training recordings of intersections of the same location that will be analyzed later in the test, but in any case a recording has been included in both training and test. In addition, it would be interesting to propose an alternative that compares temporal horizons, instead of setting the 12 prediction frames (4.8 seconds) as the only horizon.

As we can see, the Vanilla-TF model is behind AMENet and DCENet in this test, while Oriented-TF improves the results to some extent, without outperforming those described. Thus, in this test it is not possible to conclude categorically whether the inclusion of heading improves trajectory forecasting.

For this reason, in the following comparative tests of generalization of the models, different splits will be selected, depending on the type of test to be performed, which avoid the visualization of equivalent scenes by the model in the training set. Furthermore, from now on only vehicles (cars, trucks, vans, trailers, etc.) will be evaluated.

\addtolength{\tabcolsep}{4pt}

\begin{table}[t]
        \caption{General Performance}
        \vspace{-4pt}
        \label{tab:comparative-table-intersections}
        \centering
        \begin{tabular}{c|c}
        \textbf{InD} & \multicolumn{1}{c}{Average} \\ \hline
        S-LSTM & 1.88/4.47 \\
        S-GAN & 2.38/4.66 \\
        AMENet & 0.73/1.59 \\
        DCENET & 0.69/1.52 \\
        Vanilla-TF & 1.07/2.65 \\
        Oriented-TF & 1.02/2.57 \\
        \end{tabular}
        \vspace{-16pt}
\end{table}

\subsection{Testing in different datasets}
\subsubsection{Single dataset tests}
The aim of this section is to perform tests with different data splits within each dataset, in order to analyze the performance of the model for different scenes, keeping completely separate the data with which the model is trained and the test. It is also possible to compare the performance of the original system and the one that includes the heading. As shown in Table \ref{tab:single-dataset-tests}, the Oriented-TF takes advantage in the INTERACTION recordings, improving the FDE by more than one meter in all scenarios.

As can be appreciated in the table, the results of split 4 of the inD are notably weaker in all metrics. Analyzing the recordings, it is possible to think that in this intersection the network does not have other previous references, since the only one that could be similar is 3, slightly less complex and with lower occurrence in the dataset. As expected, the results in the highD are remarkably favorable, due to the strong linear component that exists in this highway dataset. Note that in the highD the authors do not provide the heading as it, so a careful selection of another included metric, the minimal distance headway (in meters), is introduced directly instead of the heading (it is not normalized in this example).

As tested in the comparative analysis against other architectures, it was also planned to carry out a data split including similar video sequences, to evaluate the performance of the two architectures in similar conditions to the ones studied in the inD. Thus, the results obtained are as expected, with a decrease of about 4 times the error in the rounD case. This may be due to the marked imbalance of the data per scenario in this dataset. In the INTERACTION, the error is also lower, but to a minor extent, and the results are slightly better for the Oriented-TF model.

\begin{table*}[]
\vspace{-5pt}
\caption{Single dataset tests}
\vspace{-5pt}
\label{tab:single-dataset-tests}
\centering
\begin{tabular}{c|c|c||c|c|c}
\textbf{Training // Test} & \textbf{\begin{tabular}[c]{@{}c@{}}Vanilla-TF\\ ADE / FDE\end{tabular}} & \textbf{\begin{tabular}[c]{@{}c@{}}Oriented-TF\\ ADE / FDE\end{tabular}} & \textbf{Training // Test} & \textbf{\begin{tabular}[c]{@{}c@{}}Vanilla-TF\\ ADE / FDE\end{tabular}} & \textbf{\begin{tabular}[c]{@{}c@{}}Oriented-TF\\ ADE / FDE\end{tabular}} \\ \hline
\textbf{inD: 123 // 4} & \textbf{7.67} / 17.22 & 7.71 / \textbf{16.83} & \textbf{rounD: 01 // 2} & \textbf{6.59 / 16.87} & 6.62 / 17.09 \\ \hline
\textbf{inD: 124 // 3} & \textbf{1.46 / 3.85} & 1.56 / 4.08 & \textbf{rounD: 02 // 1} & \textbf{6.64 / 17.04} & 6.88 / 17.53 \\ \hline
\textbf{inD: 134 // 2} & \textbf{2.80 / 7.46} & 3.47 / 9.02 & \textbf{rounD: 12 // 0} & \textbf{6.68 / 16.71} & 7.98 / 19.82 \\ \hline
\textbf{inD: 234 // 1} & 1.91 / 5.18 & \textbf{1.89 / 5.14} & \textbf{rounD: mixed} & \textbf{1.88 / 4.85} & 1.94 / 5.10\\ \hline
\textbf{inD: mixed} & 1.07 / 2.65 & \textbf{1.02 / 2.57} & \textbf{highD} & \textbf{1.19 / 2.96} & 2.20 / 3.75\\ \hline
\textbf{\begin{tabular}[c]{@{}c@{}}INT - intersection:\\ EP0-EP1-MA // GL\end{tabular}} & 2.54 / 6.95 & \textbf{2.10 / 5.66} & \textbf{\begin{tabular}[c]{@{}c@{}}INT - roundabout:\\ SR-FR-EP-OF // LN\end{tabular}} & 4.46 / 11.65 & \textbf{3.81 / 9.51} \\ \hline
\textbf{\begin{tabular}[c]{@{}c@{}}INT - intersection: \\ MA-GL-EP0 // EP1\end{tabular}} & 3.27 / 8.17 & \textbf{2.80 / 7.16} & \textbf{\begin{tabular}[c]{@{}c@{}}INT - roundabout:\\ LN-SR-FT-EP // OF\end{tabular}} & 4.27 / 11.63 & \textbf{3.68 / 10.11} \\ \hline
\textbf{\begin{tabular}[c]{@{}c@{}}INT - intersection:\\ mixed\end{tabular}} & 2.09 / 5.85 & \textbf{1.81 / 4.98} & \textbf{\begin{tabular}[c]{@{}c@{}}INT - roundabout:\\ mixed\end{tabular}} & 2.75 / 7.78 & \textbf{2.31 / 6.38} 
\end{tabular}
\end{table*}

\subsubsection{Mixing datasets: similar scenarios of different datasets}
In order to assess the generalization potential of the system in terms of coordinate prediction independently of the inputs, this analysis will test the model in similar scenarios to those already known, but from a completely different dataset. 

As can be seen in table \ref{tab:equivalent-scenario-tests}, the results are quite satisfactory for the intersections, obtaining similar figures to those obtained by performing the train on the dataset itself. Something specific can be observed in the case of roundabouts, where a lower error is obtained when testing on roundabouts of a dataset different from the dataset on which the model has been trained. In the case of the Oriented-TF, no improvement is seen here, being marginal only for the highD.
The generalization of the model in this case is excellent, obtaining results that are even better than its own.

\begin{table}[]
\vspace{-4pt}
\caption{Equivalent scenario tests (training on entire dataset)}
\label{tab:equivalent-scenario-tests}
\vspace{-6pt}
\centering
\setlength\extrarowheight{-0.5pt}
\begin{tabular}{c|c|c}
\textbf{Training // Test} & \textbf{\begin{tabular}[c]{@{}c@{}}Vanilla-TF\\ ADE / FDE\end{tabular}} & \textbf{\begin{tabular}[c]{@{}c@{}}Oriented-TF\\ ADE / FDE\end{tabular}} \\ \hline
\textbf{\begin{tabular}[c]{@{}c@{}}inD // INT-int\end{tabular}} & \textbf{3.12 / 8.10}  & 4.89 / 10.87 \\ \hline
\textbf{\begin{tabular}[c]{@{}c@{}}INT-int // inD\end{tabular}} & \textbf{4.04 / 10.10} &  4.24 / 10.32 \\ \hline
\textbf{\begin{tabular}[c]{@{}c@{}}rounD // INT-round\end{tabular}} & \textbf{3.19 / 8.34} & 5.18 / 11.72 \\ \hline
\textbf{\begin{tabular}[c]{@{}c@{}}INT-round //  rounD\end{tabular}} & \textbf{5.30 / 14.13} & 6.99 / 16.54 \\ \hline
\textbf{\begin{tabular}[c]{@{}c@{}}highD // INT-merg\end{tabular}} & 2.45 / 5.14 & \textbf{2.35 / 4.77} \\
\end{tabular}
\vspace{0pt}
\end{table}

\subsubsection{Generalization between different scenarios}
After testing comparable scenarios, the performance of vehicle dynamics learning will now be assessed, independently of the trajectories observed in the training videos. So, for example, could it know how a vehicle will act at a junction if it has been trained with roundabouts? 
As observed in Table \ref{tab:different-scenario-tests}, the generalization in this case is also fairly adequate, highlighting an improvement to the single results in the inD-rounD and inD-INT-round test. It seems quite significant that the model has improved the results in roundabouts training with intersections, and it is also remarkably the performance improvement of the Oriented-TF in the training and test cases in the INTERACTION.

\begin{table}[t]
\vspace{-1pt}
\caption{Different scenarios tests (training on entire dataset)}
\label{tab:different-scenario-tests}
\centering
\setlength\extrarowheight{-0.5pt}
\begin{tabular}{c|c|c}
\textbf{Training // Test} & \textbf{\begin{tabular}[c]{@{}c@{}}Vanilla-TF\\ ADE / FDE\end{tabular}} & \textbf{\begin{tabular}[c]{@{}c@{}}Oriented-TF\\ ADE / FDE\end{tabular}} \\ \hline
\textbf{\begin{tabular}[c]{@{}c@{}}inD // rounD\end{tabular}} & \textbf{5.87 / 15.08} & 5.97 / 15.37 \\ \hline
\textbf{\begin{tabular}[c]{@{}c@{}}rounD // inD\end{tabular}} & \textbf{3.27 / 8.35} & 3.40 / 8.59 \\ \hline
\textbf{\begin{tabular}[c]{@{}c@{}}INT-int // INT-round\end{tabular}} & 5.04 / 12.84 &  \textbf{4.51 / 11.68}\\ \hline
\textbf{\begin{tabular}[c]{@{}c@{}}INT-round // INT-int\end{tabular}} & 2.99 / 8.21 & \textbf{2.67 / 7.28} \\ \hline
\textbf{\begin{tabular}[c]{@{}c@{}}inD // INT-round\end{tabular}} & \textbf{3.34 / 8.58} & 5.26 / 11.46 \\ \hline
\textbf{\begin{tabular}[c]{@{}c@{}}INT-round // inD\end{tabular}} & \textbf{3.36 / 8.83} & 4.44 / 10.08
\end{tabular}
\vspace{-12pt}
\end{table}

\subsubsection{Changing frame rate of input data - Vanilla TF}
The frequency of data input also seems to be vital in the performance of a prediction system. In this section, Vanilla-TF has been tested on the set of inD vehicles, with training strategy on 3 scenes and test on the remaining one. The tests have been carried out with the original layout of 2.5 fps (8-12) and on 5 fps (16-24).
For the test sets on scenarios 1 and 2, the choice of doubling the framerate is the winning option, with FDE improvements of 0.43 m and 0.57 m, respectively. However, for scenarios 3 and 4, the 2.5 fps framerate is still the better option, with FDE improvements in favor of 1.49 m and 3.5 m, respectively. Junctions 1 and 2 coincide with crossroads at perpendicular intersections where the vehicle dynamics are different to the others, possibly extracting more information and taking advantage of the additional framerate.

\subsection{Qualitative results}
\begin{figure*}[t]
\centering
\captionsetup{justification=centering}
\captionsetup{position=top}
\subfloat[Linear trajectory.]{
    \includegraphics[clip, width=0.305\textwidth,valign=c]{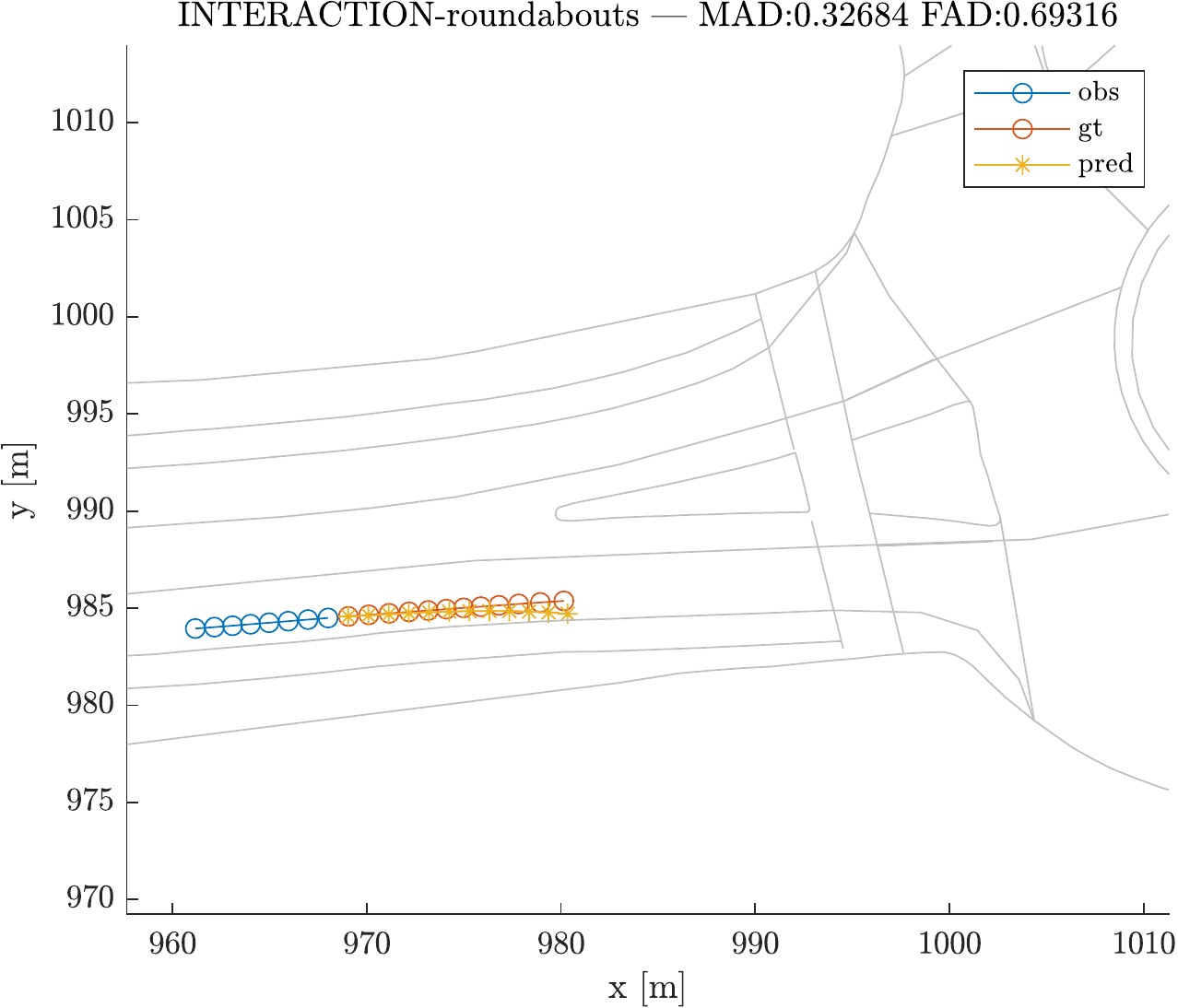}
            }
\subfloat[Turn - good prediction.]{
    \includegraphics[clip, width=0.322\textwidth,valign=c]{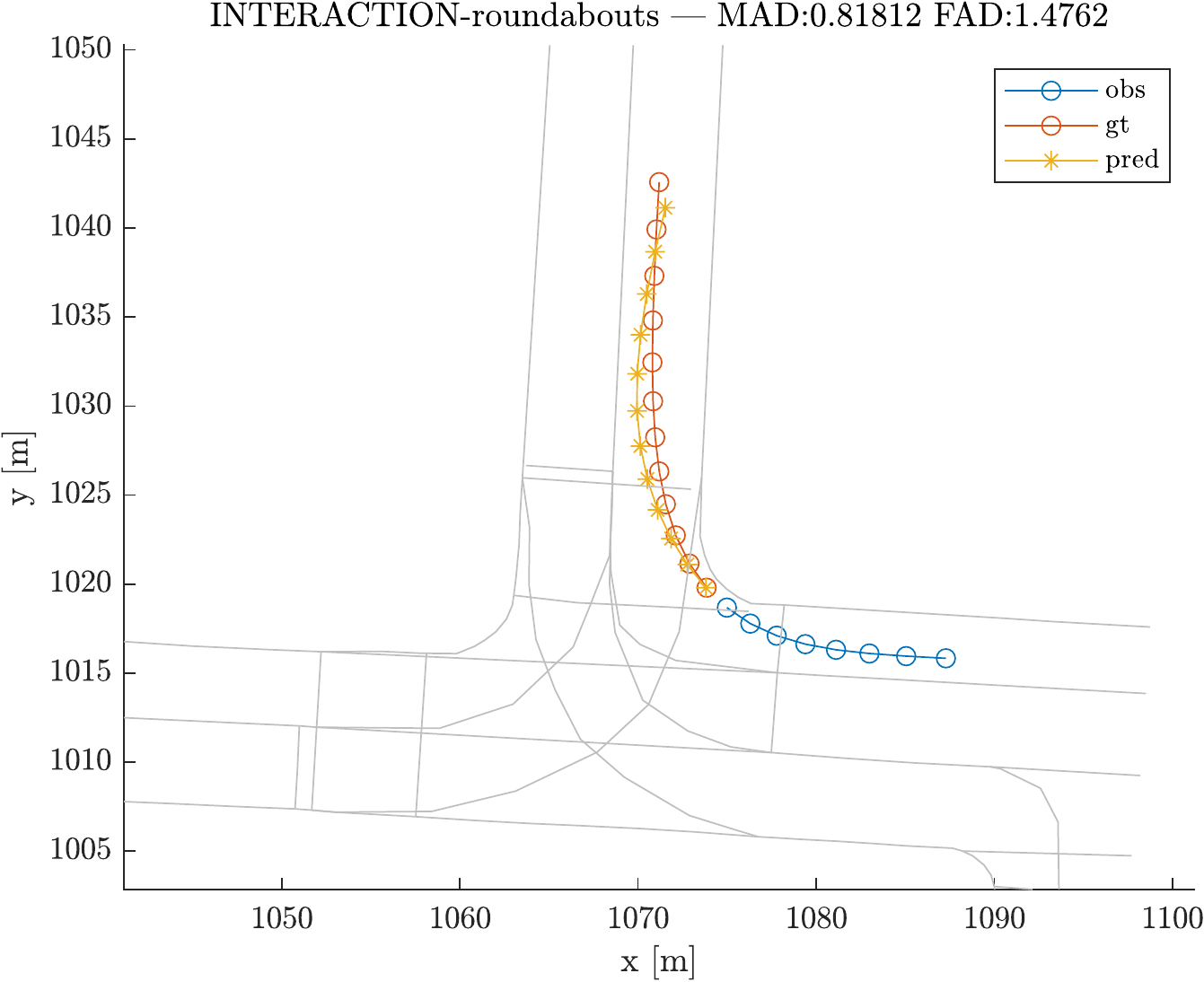}
            }
\subfloat[Wrong prediction.]{
    \includegraphics[clip, width=0.322\textwidth,valign=c]{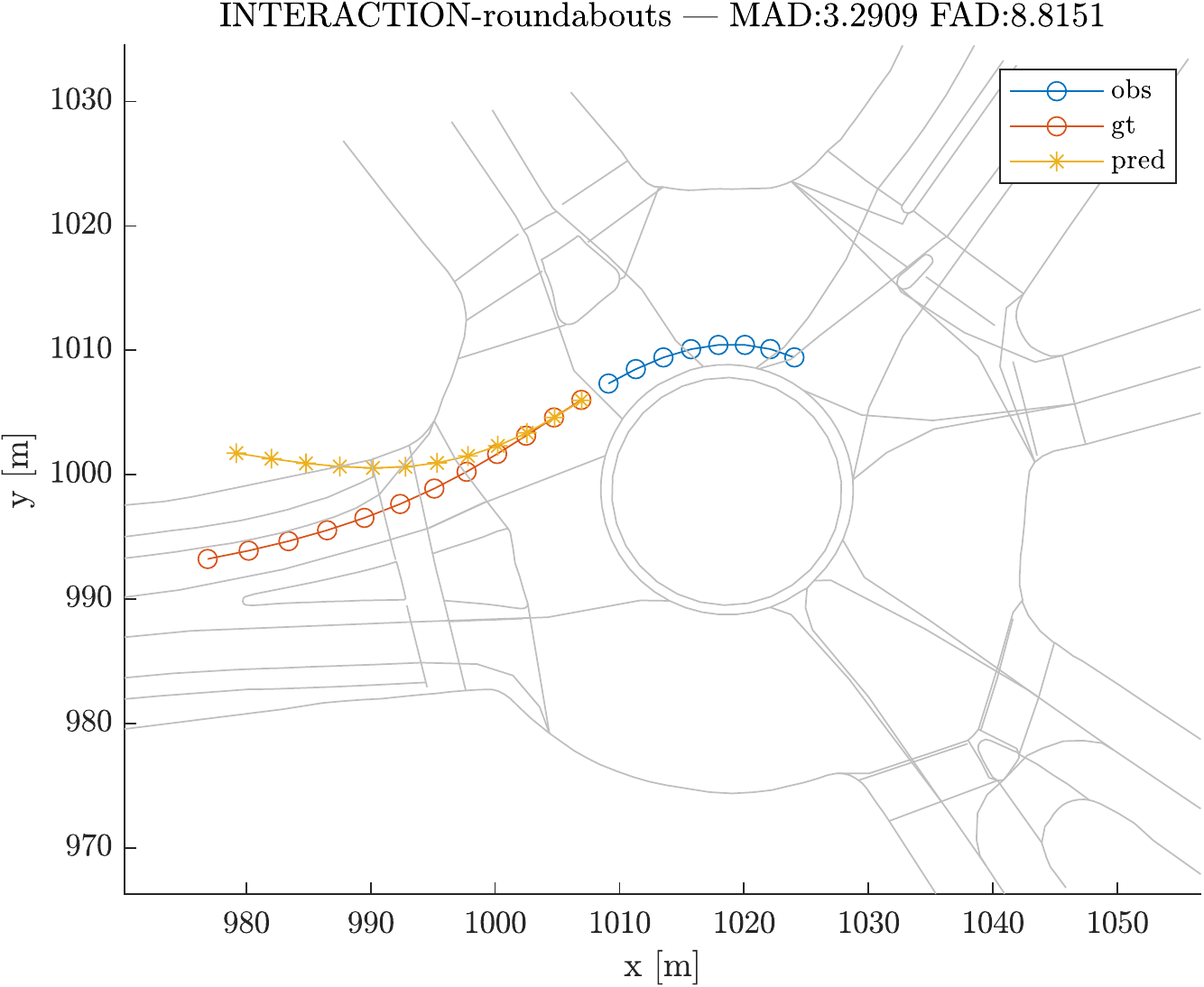}
            }
\vspace{-4pt}
\caption{Sample outputs for cross experiment (trained in rounD - tested in INTERACTION Roundabouts set). Observed trajectory is depicted in blue, ground truth in red and predicted trajectory in yellow.}
\label{fig:sample-outputs}
\end{figure*}

\begin{figure}[t]
    \centering
    {
    \vspace{2pt}
    \includegraphics[width=0.89\columnwidth]{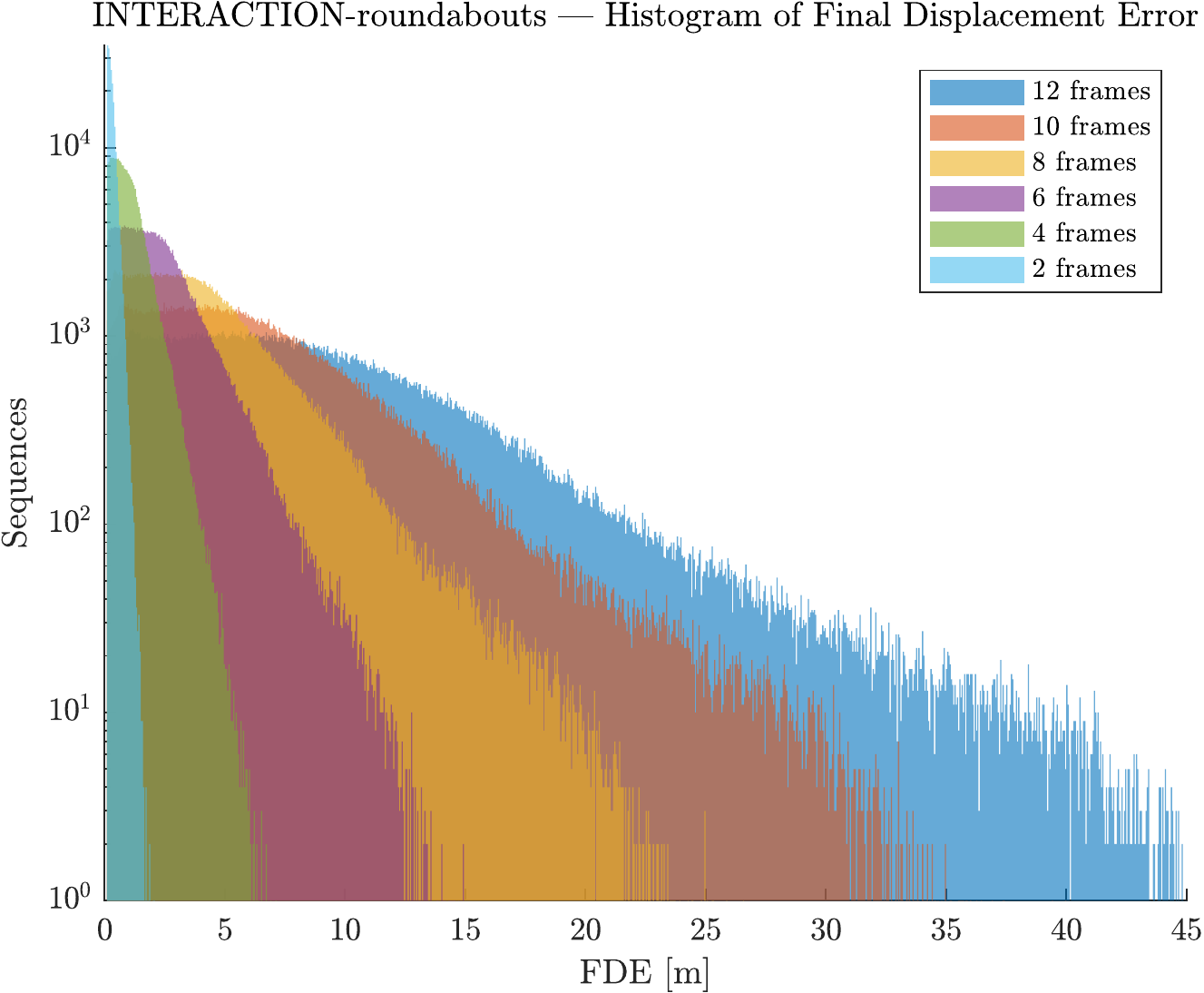}
    }
    \vspace{-9pt}
    \caption{Histogram of FDE.}
    \vspace{-16pt}
    \label{fig:histogramFDE}
\end{figure}

Beyond the quantitative results, it is always convenient to have an approach closer to reality by directly representing the input and output data of the system. Figure \ref{fig:sample-outputs} shows three prediction situations that were observed in one of the cross experiments, specifically in the rounD - INTERACTION Roundabouts. In one of them the system has correctly predicted a linear trajectory, in another it is forecasting quite correctly a moderately tight turn, while in the last one it has chosen a turn in the wrong direction, making a very significant error according to the established metrics. The selected figures include one of the sections of the USA\_EP map, where there is a junction stretch adjacent to the roundabout. Thus, it is also possible to appreciate the model's generalization capacity, which has been solely trained with European roundabouts from the rounD dataset. 
The figure \ref{fig:histogramFDE} shows the histogram of FDE for various forecast horizons in the qualitative scenario, considering in all cases 8 frames observed. It is visible that the errors increase as the time horizon is extended, showing slopes similar to those of a normal distribution.

As seen in this section, BEV datasets used with the TF can deliver surprising results, performing better in some situations when they have been trained with foreign scenes. This raises the question of whether these scenarios really contribute anything to model learning, opening the debate as to whether it is better to have a large amount of data or whether more variability in the scenes, coupled with detailed labeling, is more valuable.
An interesting phenomenon has also been noticed with the inclusion of heading, working better in the INTERACTION dataset, while in the others it has hardly improved. This opens the door to a modification of the normalization method and processing of heading in these datasets.

\section{Conclusions and Future Work} \label{sec:conclusions}
Based on the experiments performed, it is possible to conclude that the Oriented-TF model, as well as Vanilla-TF, are fully competent among the state-of-the-art models for the datasets analyzed in this work, confirming its good performance in TrajNet by its original authors, considering that it is a single agent approach, where no context variables or interaction with other agents are included. Thus, a first approach to the analysis of its generalization ability has also been carried out, by conducting multiple cross-tests between similar scenarios of diverse datasets, analyzing the obtained results. A novel use of the Transformer is also proposed, by adding the agent's orientation as an input variable to improve the trajectory prediction, observing interesting results, depending on the dataset analyzed
As future work, the core task is to further develop the model, measuring its possibilities for single agent input data processing, as well as exploring the social architectures already proposed based on graphs. In addition, the direct use of this model could involve other datasets that also contain data that can be expressed in 2D, as is the case of the information that we can obtain from PREVENTION through the radars. This will enable testing, for example, the inference time in a real situation by obtaining information from the radars of an instrumented vehicle.
\vspace{-4pt}

\section*{ACKNOWLEDGMENT}
\vspace{-4pt}This work was funded by Research Grants S2018/EMT-4362  (Community Reg. Madrid), DPI2017-90035-R  (Spanish Min. of Science and Innovation), BRAVE Project, H2020, Contract \#723021 and PRE2018-084256 (Spanish Min. of Education) via a predoctoral grant to the first author. 

%


\bibliographystyle{bibliography/IEEEtran}
\bibliography{bibliography/IEEEabrv,bibliography/references }

\end{document}